\documentclass[sigconf,screen]{acmart}

\usepackage{booktabs} 
\usepackage{amsfonts}
\usepackage{amssymb}
\usepackage{xspace}
\usepackage{epstopdf}
\usepackage{enumitem}
\usepackage{algorithm}
\usepackage{algorithmic}
\usepackage{multirow}
\usepackage{balance}
\usepackage{booktabs}
\usepackage{array}
\usepackage{wrapfig}
\usepackage{pifont}
\usepackage{pbox}
\usepackage{geometry}
\usepackage{color}
\setcopyright{none}
\renewcommand\footnotetextcopyrightpermission[1]{} 
\thanks{Accepted for publication in the Proceedings of 20th ACM International Conference on Multimodal Interaction, Boulder, CO, USA.}
\thanks{DOI: 10.1145/3242969.3242988}

\begin{document}


\def\x{{\mathbf x}}
\def\L{{\cal L}}
\def\x{{\mathbf x}}
\def\L{{\cal L}}
\def\eg{\textit{e.g.}}
\def\ie{\textit{i.e.}}
\def\Eg{\textit{E.g.}}
\def\etal{\textit{et al.}}
\def\etc{\textit{etc.}}
\def\nback{\textit{n}-back }
\setlength{\tabcolsep}{2pt}

\title{Looking Beyond a Clever Narrative: Visual Context and Attention are Primary Drivers of Affect in Video Advertisements}

\author{Abhinav Shukla}
\affiliation{%
  \institution{International Institute of Information Technology}
  \city{Hyderabad} 
  \country{India} 
}
\email{abhinav.shukla@research.iiit.ac.in}

\author{Harish Katti}
\affiliation{%
  \institution{Centre for Neuroscience, Indian Institute of Science}
  \streetaddress{Old TIFR building, Gulmohar Marg, Mathikere}
  \city{Bangalore} 
  \country{India}
	\postcode{560012}}
\email{harish2006@gmail.com}

\author{Mohan Kankanhalli} 
\affiliation{%
 \institution{School of Computing, National University of Singapore}
 \streetaddress{21, Lower Kent Ridge Road}
 \city{Singapore} 
 \postcode{117417}
 \country{Singapore}}
\email{mohan@comp.nus.edu.sg}

\author{Ramanathan Subramanian}
\affiliation{%
  \institution{Advanced Digital Sciences Center, University of Illinois}
  \streetaddress{1 Create Way, \#14-02
Create Tower}
  \city{Singapore} 
    \country{Singapore}
}
\email{ramanathan.subramanian@ieee.org}

\renewcommand{\shortauthors}{A. Shukla, H. Katti, M. Kankanhalli, and R. Subramanian}
\renewcommand{\shorttitle}{Visual Context and Attention are Primary Drivers of Affect in Video Advertisements.}

\begin{abstract}
Emotion evoked by an advertisement plays a key role in influencing brand recall and eventual consumer choices. Automatic ad affect recognition has several useful applications. However, the use of content-based feature representations does not give insights into how affect is modulated by aspects such as the ad scene setting, salient object attributes and their interactions. Neither do such approaches inform us on how humans prioritize visual information for ad understanding. Our work addresses these lacunae by decomposing video content into detected objects, coarse scene structure, object statistics and actively attended objects identified via eye-gaze. We measure the importance of each of these information channels by systematically incorporating related information into ad affect prediction models. Contrary to the popular notion that ad affect hinges on the narrative and the clever use of linguistic and social cues, we find that actively attended objects and the coarse scene structure better encode affective information as compared to individual scene objects or conspicuous background elements.
\end{abstract}


\maketitle

\section{Introduction}~\label{sec:intro}
We encounter video-based advertising on a daily basis on television and websites like YouTube\textregistered. Advertising is a multi-billion dollar industry worldwide that seeks to capture user attention and maximize short and long term brand recall, ad monetization and click-based user interactions \cite{Farris2010}. A prevalent notion in advertisement (ad) design~\citep{Adriana2018} is that the ad narrative is key~\cite{White2006}. From an analysis viewpoint, narrative analytics would require one to parse out the story line, subtle forms of communication such as motivation~\cite{Hussain2017}, sarcasm, humor and sentiments in an ad~\cite{Adriana2018}. Such data often does not accompany video content in popular annotated databases~\citep{baveye2015liris}, and are often difficult to analyze reliably.\\
Contrary to this view, advertisement evaluation and impact~\cite{Broach1995,Kamins1991} have been attributed to evoked emotions of pleasantness (valence) and arousal~\cite{Truss132}. Past research has also shown that affect-based video ad analytics compares favorably~\cite{cavva} to visual similarity and text-based analysis~\cite{videosense}. Affect based content analysis works well in combination with behavioural read-outs of cognitive load and arousal~\cite{YadatiMMM2013} as well as user attention~\cite{Shukla2017icmi}. Computational work on affect-based ad analysis has shown that analyzing even 10 second segments can give meaningful prediction of induced emotional states~\cite{Shukla2017acm,Shukla2017icmi,cavva}. Overall, affect based analysis seems much more robust than semantic based approaches which would require the narrative of nearly the entire ad for a reliable evaluation to be made. 

Ad affect prediction becomes important in the context of computational advertising as online ads are often interrupted or skipped by users as they have little similarity/relevance to the streamed videos within which they are embedded. Despite the success of content based affect prediction, there is little or no understanding of which aspects of ad video information best inform the induced valence and arousal. For example, is the setting or mood of an advertisement more important than objects that are part of the story line? Are all objects equally important? Lack of knowledge on these questions is a conceptual bottle-neck to design principled methods for video advertisement analysis.\\
In this paper, we set out to address this exact question by parceling out video advertisement content into independent channels capturing various aspects of the scene content. Exemplar channels include (i) the coarse scene structure that captures the overall mood or ambiance of the video scene, (ii) collection of the (detected) scene objects, (iii) scene objects that are interesting from the user's perspective, and (iv) features encoding the list of scene objects and their co-occurrence (object context). Scene objects capturing overt visual attention during a given time interval in the video~\cite{retar2013} are inferred from eye-gaze recordings of users viewing ad videos. 

These channels have the added advantage that they are easily interpretable, and insights derived therefrom can be incorporated into computational frameworks as well as the ad design. This method of parceling media content into distinct channels is inspired from recent approaches in visual neuroscience~\citep{katti2017targets,katti2016object,groen2018distinct} to examine various behavioral and neural read outs of human perception. We then train predictive models for valence and arousal upon processing each of these visual channels. Analysis of these models not only conveys the effectiveness of each of these channels in predicting affect; our experiments reveal the unexpected and critical role played by the coarse grained scene structure in conveying valence and arousal.

The main contributions of our paper are:-
\begin{enumerate}
\item We demonstrate a novel approach for advertisement video analysis by decomposing video information into multiple, independent information channels.
\item We show that a model incorporating information from both the user and scene content best predicts the ad emotion.
\item We bring forth the critical role played by the coarse scene structure towards conveying the ad affect.
\item We crystallize insights from our analysis to propose rules suitable for the design and creation of video ads, and computational advertising frameworks.
\end{enumerate}

Our paper is organized as follows. Section \ref{sec:rw} discusses related work and espouses the novelty of our work with respect to existing literature. Section \ref{sec:dataset} describes the ad dataset used in our experiments. Section \ref{sec:methods} describes the various video channels created from the original ad content, while Section \ref{sec:er} discusses experiments and related findings. Section \ref{sec:cfw} concludes the paper.

\section{Related Work}~\label{sec:rw}
To position our work with respect to existing literature and to establish its novelty, we examine related work on: (i) Advertisements and Emotions, (ii) Visual Content Analysis (especially for affect recognition and understanding scene structure and semantics) and (iii) Human Centered Affect Recognition.

\subsection{Advertisements and Emotions}
Advertisements are inherently emotional, and have long been shown to modulate user behaviour towards products in response to the emotions they induce~\cite{Holbrook1984,Holbrook1987,Pham2013}. Holbrook and Batra~\cite{Holbrook1987} observe correlations between emotions induced by ads, and  users' brand attitude. Pham \etal~\cite{Pham2013} note that ad-evoked feelings impact users both explicitly and implicitly. Recent work has aimed at exploiting the affective information present in ads for computational advertising, \ie, inserting the most emotionally relevant ads in scene transition points within programs. Shukla \etal~\cite{Shukla2017acm,Shukla2017icmi} show that improved audiovisual affect recognition can enhance ad recall and improve viewer experience when inserting ads in videos. Products deployed in the industry such as \textit{Creative Testing} and \textit{Media Planning} by RealEyes \cite{realeyes}, which provide second-by-second emotion traces and predict exactly how people react to media content, thereby revealing what works and where the pitfalls lie. These methods exploit emotional content and structure of advertisements to perform precise ad characterization, which underlines the importance of emotions in ads.

\subsection{Visual Content Analysis for Affect Recognition}

A very important aspect of our work is understanding the importance of various visual scene elements for affect prediction. Classical works in visual affect recognition such as \cite{Hanjalic2005,pantic2009survey,gunes2010automatic} focus on low level handcrafted features such as motion activity, color histograms \etc~Recent video analytic works have leveraged the advances in deep learning~\cite{Shukla2017acm,morency2017icmi}. Combined with the breakthroughs in deep learning since AlexNet \cite{alex12}, the Places database \cite{places14} enables training of deep learning models for scene attribute recognition on a large scale. There have also been notable community challenges \cite{ringeval2017avec,dhall2017emotiw} that focus especially on affect recognition from visual content analysis in various settings. Although all of these works focus on affect prediction from content, they do not try and measure the affective content encoded by the various scene elements. Attempts to decompose visual information and study user behavioural measures have risen lately. Katti \etal~\cite{katti2017targets,katti2016object} have explored human response times to object detection and human beliefs for object presence from various kinds of visual information. Groen \etal~\cite{groen2018distinct} have evaluated the distinct contributions of functional and deep neural network features to representational similarity of scenes in human brain and behaviour. Karpathy \etal~\cite{karpathy2014large} utilize a foveated stream and a low resolution context stream for large scale video classification. Xu \etal~\cite{xu2015show} utilize a neural network to model human visual attention (computerized gaze) for image caption generation. Fan \etal~\cite{fan2018attention} use eye tracking as well as saliency modeling to study how image sentiment and visual attention are linked. Our work builds on the foundations laid by \cite{karpathy2014large,fan2018attention,xu2015show} and specifically analyzes the contributions of various scene elements and channels towards capturing affective information in the confined field of advertisement videos.

\subsection{Human Centered Affect Recognition}
Human physiological signals have been widely employed for implicit tagging of multimedia content in recent years. Implicit tagging has been defined as using non-verbal spontaneous behavioral responses to find relevant tags for multimedia content \cite{pantic2009implicit}. These responses include Electroencephalography (EEG) and Magnetoencephalography (MEG) along with physiological responses such as heart rate and body temperature~\cite{koelstra2009eeg,Shukla2017icmi,Abadi13,Wache2015,soleymani2012mahnob} and gazing behavior~\cite{soleymani2012multimodal,Ramanathan2009,Gilani15,Viral18}. Determining the affective dimensions of valence and arousal using these methods has been shown to be very effective. Affective attributes have also been shown to be useful for retrieval \cite{tkalvcivc2010using}. These methods have an advantage of not requiring additional human effort in annotation, and are based on the analysis of spontaneous user reactions or behavioral signals. We employ eye gaze as a modality of interest in our experiments.

\subsection{Analysis of Related Work}
On examining the related work, we find that: (1) Even though emotions are very important in visual advertising, there have rarely been attempts to computationally analyze how they are represented in the scene structure and setting, (2) Characterizing affective visual information has been achieved primarily via computational cues alone and not by human driven cues, (3) Despite the fact that decomposition of visual information has proved to be useful for determining user behavioural traits, there has been no such work in the context of affect recognition.

In this regard, we present the first work to study the contribution of different types of visual information to affect recognition and combine content with human driven information from eye gaze. We evaluate the performance of features extracted from various information channels on an affective dataset of advertisements described below.

\section{Dataset}~\label{sec:dataset}
This section presents details regarding the ad dataset used in this study along with the protocol employed for collecting eye tracking data, which is used in some affect recognition experiments.

\subsection{Dataset Description}
\begin{figure}[t]
\includegraphics[width=0.32\linewidth]{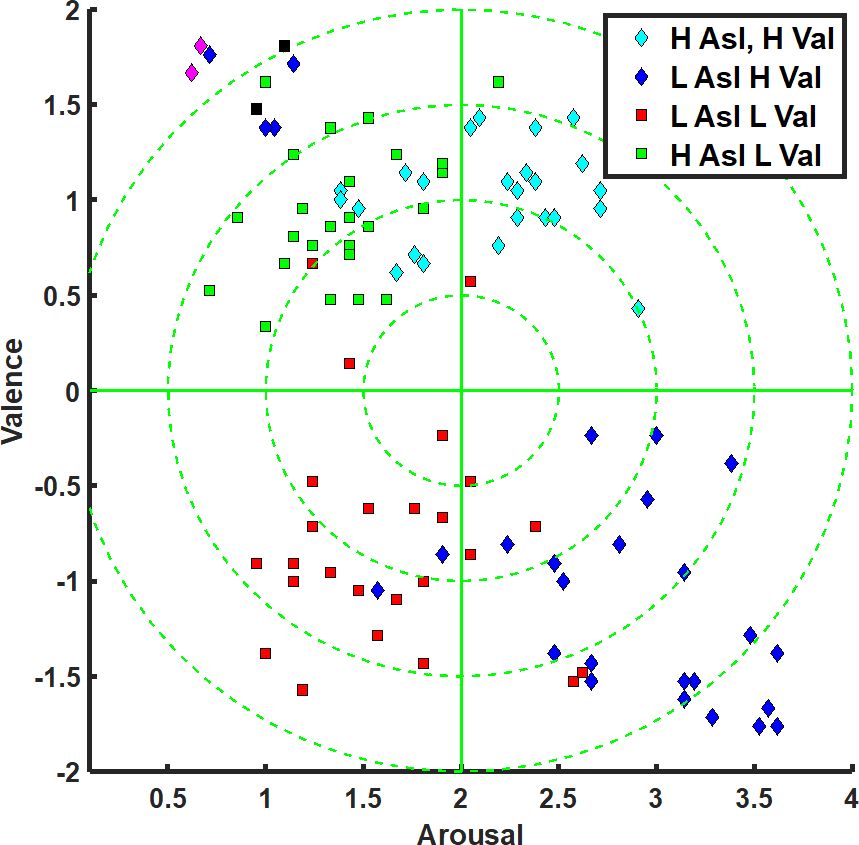}\hfill
\includegraphics[width=0.33\linewidth]{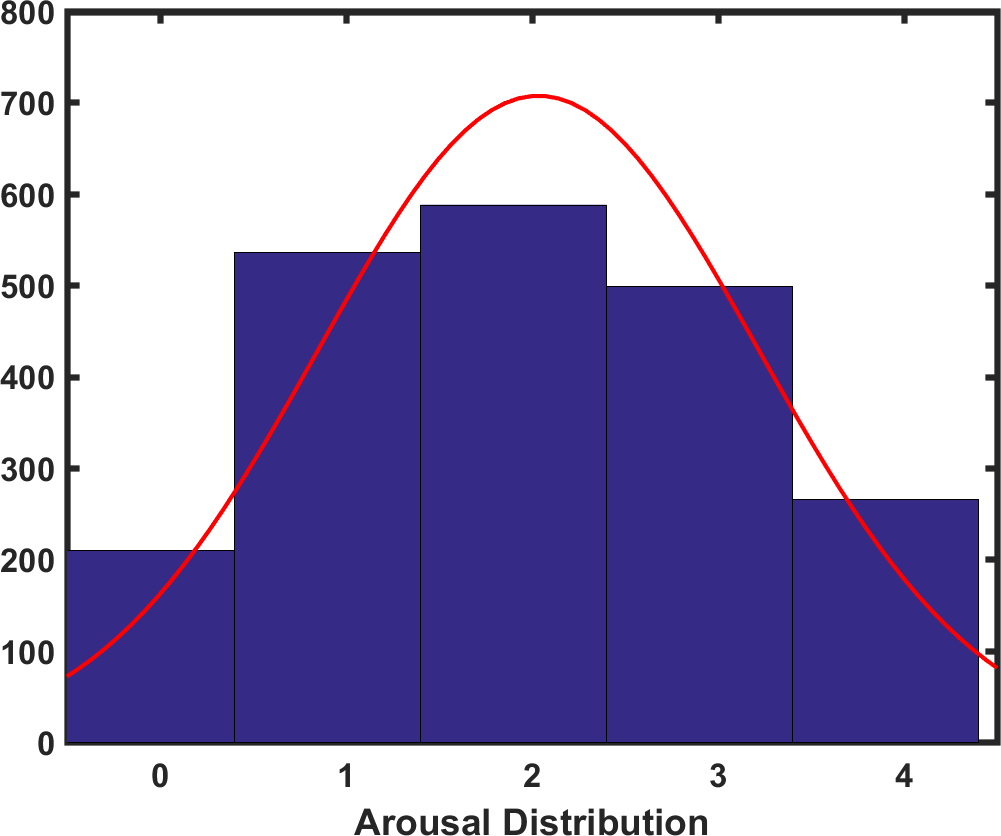}\hfill
\includegraphics[width=0.33\linewidth]{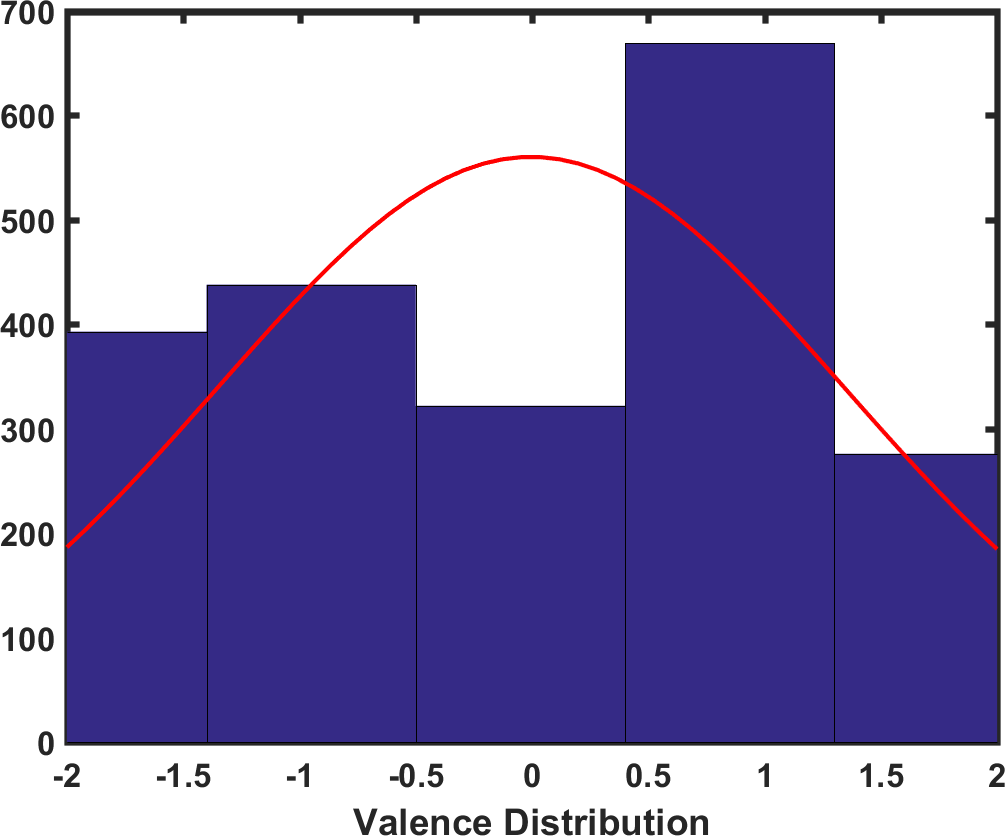}\vspace{-.1cm}
\caption{\label{Annot_dist} (left) Scatter plot of mean arousal, valence ratings obtained from 23 users. Color-coding based on expert labels. (middle) arousal and (right) valence score distributions (view under zoom).}\vspace{-.3cm}
\end{figure}

Defining \textbf{\textit{valence}} as the feeling of \textit{pleasantness}/\textit{unpleasantness} and \textbf{\textit{arousal}} as the \textit{intensity of emotional feeling} while viewing an advertisement, five experts compiled a dataset of 100, roughly 1-minute long commercial advertisements (ads) in~\cite{Shukla2017acm}. These ads are publicly available\footnote{On video hosting websites such as YouTube} and roughly uniformly distributed over the arousal--valence plane defined by Greenwald \textit{et al.}~\cite{greenwald1989} (Figure~\ref{Annot_dist}(left)) as per the first impressions of 23 naive annotators. The authors of~\cite{Shukla2017acm} chose the ads based on consensus among five experts on valence and arousal labels (either \textit{high} (H)/\textit{low} (L)). Labels provided by ~\cite{Shukla2017acm}  were considered as \textbf{\textit{ground-truth}}, and used for affect recognition experiments in this work. 
 
To evaluate the effectiveness of these ads as affective control stimuli, we examined how consistently they could evoke target emotions across viewers. To this end, the ads were independently rated by 23 annotators for valence and arousal in this work\footnote{~\cite{Shukla2017acm} employs 14 annotators.}. All ads were rated on a 5-point scale, which ranged from -2 (\textit{very unpleasant}) to 2 (\textit{very pleasant}) for valence and 0 (\textit{calm}) to 4 (\textit{highly aroused}) for arousal. Table~\ref{tab:ads_des} presents summary statistics for ads over the four quadrants. Evidently, low valence ads are longer and are perceived as more arousing than high valence ads, implying that they had a stronger emotional influence among viewers.

We computed agreement among raters in terms of the (i) Krippendorff's $\alpha$ and (ii) Fleiss $\kappa$ scores. The $\alpha$ coefficient is applicable when multiple raters supply ordinal scores to items. We obtained $\alpha = 0.62$ and $0.36$ respectively for valence and arousal, implying that valence impressions were more consistent across raters. On a coarse-grained scale, we then computed the Fleiss $\kappa$ agreement across annotators-- the Fleiss Kappa statistic is a generalization of Kohen's Kappa, and is applicable when multiple raters assign categorical (\textit{high}/\textit{low} in our case) to a fixed number of items. Upon thresholding each rater's arousal, valence scores by their mean rating to assign \textit{high}/\textit{low} labels for each ad, we observed a Fleiss $\kappa$ of 0.56 (moderate agreement) for valence and 0.27 (fair agreement) for arousal across raters. Computing Fleiss $\kappa$ upon thresholding each rater's scores with respect to the grand mean, we obtained a Fleiss $\kappa$ of 0.64 (substantial agreement) for valence, and 0.30 (fair agreement) for arousal. Clearly, there is reasonable agreement among the novice raters on affective impressions with considerably higher concordance for valence. These results led us to confirm that the 100 ads compiled by ~\cite{Shukla2017acm} are appropriate for affective studies.  

To examine if the arousal and valence dimensions are independent based on user ratings, we (i) examined scatter plots of the annotator ratings, and (ii) computed correlations amongst those ratings. The scatter plot of the mean arousal, valence annotator ratings, and the distribution of arousal, valence ratings are presented in Fig.~\ref{Annot_dist}. The scatter plot is color-coded based on expert labels, and is more spread out than the classical 'C' shape observed with affective datasets~\cite{IAPS}, music videos~\cite{Koelstra} and movie clips~\cite{decaf}. Arousal and valence scores are spread across the rating scale, with a greater variance observable for valence ratings. Wilcoxon rank sum tests on annotator ratings revealed significantly different mean arousal ratings for high and low arousal ads ($p<0.0001$), and distinctive mean valence scores for high and low valence ads ($p<0.0001$), consistent with expectation.    

Pearson correlation was computed between the arousal and valence dimensions by limiting the false discovery rate to within 5\%~\cite{benjamini1995controlling}. This procedure revealed a weak and insignificant negative correlation of 0.16, implying that ad arousal and valence scores were largely uncorrelated. Overall, observed results suggest that (a) the compiled ads evoked a wide range of emotions across the arousal and valence dimensions, and (b) there was considerable concordance among annotator impressions especially for valence, and they provided distinctive ratings for emotionally dissimilar ads.

\subsection{Eye Tracking Data Acquisition Protocol}~\label{eyetrackingprotocol}
As the annotators rated the ads for arousal and valence, we acquired their eye tracking data using the \textit{Tobii EyeX} eye tracker. The eye tracker has a sampling rate of 60 Hz and an advertised operating range of 50-90~cm. It was placed in a fixed position under the screen on which the ads were presented. Before each recording session, the eye tracker was calibrated using its sample stimulus presentation for each subject. To minimize tracking errors and fatigue effects, we allowed each annotator to take comfort breaks after viewing 20 ads. Before they began with the next set of 20 ads, we recalibrated the eye tracker to ensure signal quality. We presented the videos on an LED screen with a resolution of $1366\times768$ pixels. Each ad was preceded by a 1s fixation cross to orient user attention. We recorded a raw stream of the \textit{x} and \textit{y} pixel coordinates of the gaze, along with the gaze timestamps in milliseconds.

\begin{table}[t]
\fontsize{7}{7}\selectfont
\renewcommand{\arraystretch}{1.8}
\caption{\label{tab:ads_des} Summary statistics compiled from 23 users.} \vspace{-.2cm}
\centering
\begin{tabular}{|c|ccc|} \hline
\textbf{Quadrant} & \textbf{Mean length (s)} & \textbf{Mean arousal} & \textbf{Mean valence}  \\ \hline \hline
\textbf{H arousal, H valence} & 48.16 & 2.19 & \ 0.98 \\
\textbf{L arousal, H valence} & 44.18 & 1.47 & \ 0.89 \\
\textbf{L arousal, L valence} & 60.24 & 1.72 & -0.74 \\
\textbf{H arousal, L valence} & 64.16 & 2.95 & -1.10 \\ \hline
\end{tabular}
\vspace{-.3cm}
 \end{table}

\section{Visual Channels for Affect Recognition}~\label{sec:methods}
This section discusses the various channels of visual information that we employed for affect recognition including content-driven and human-driven (eye gaze-based). For ad affect recognition, we represent the {visual} modality using representative frame images following~\cite{Shukla2017acm}. We fine-tuned \textit{Places205} via the LIRIS-ACCEDE~\cite{baveye2015liris} dataset, and used the penultimate fully connected layer (FC7) descriptors for our analysis. From each video in the ad and LIRIS-ACCEDE datasets, we sample one frame every three seconds which enables extraction of a continuous video profile for affect recognition. This process generated 1793 representative frames for the 100 advertisement videos.

\subsection{Visual Channels and Features}
A description of each visual channel and type of feature employed for affect recognition follows. The method title contains in parentheses an abbreviation for easier reference in subsequent sections.

\subsubsection{Raw Frames (\textbf{Video})}
\begin{figure}[h]
\centering
\includegraphics[width=0.47\linewidth]{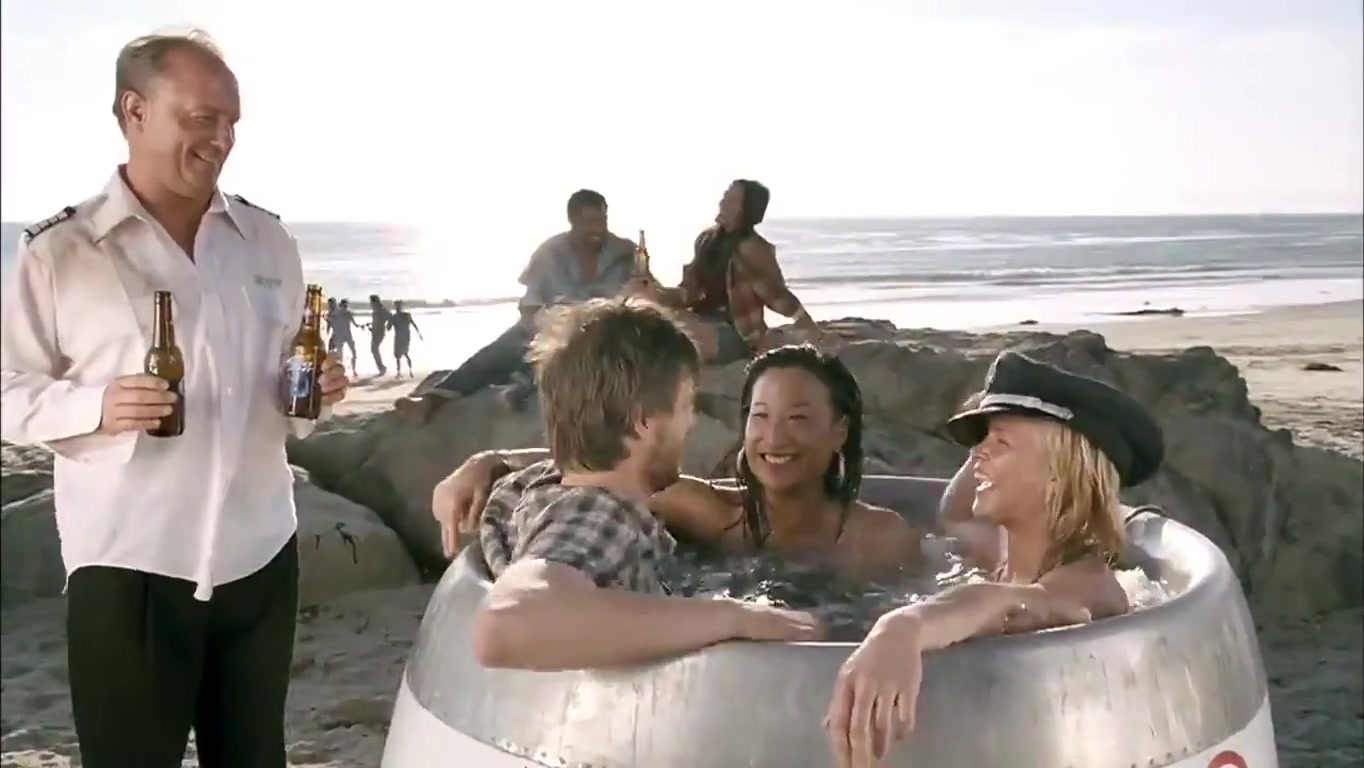}\hfill
\includegraphics[width=0.47\linewidth]{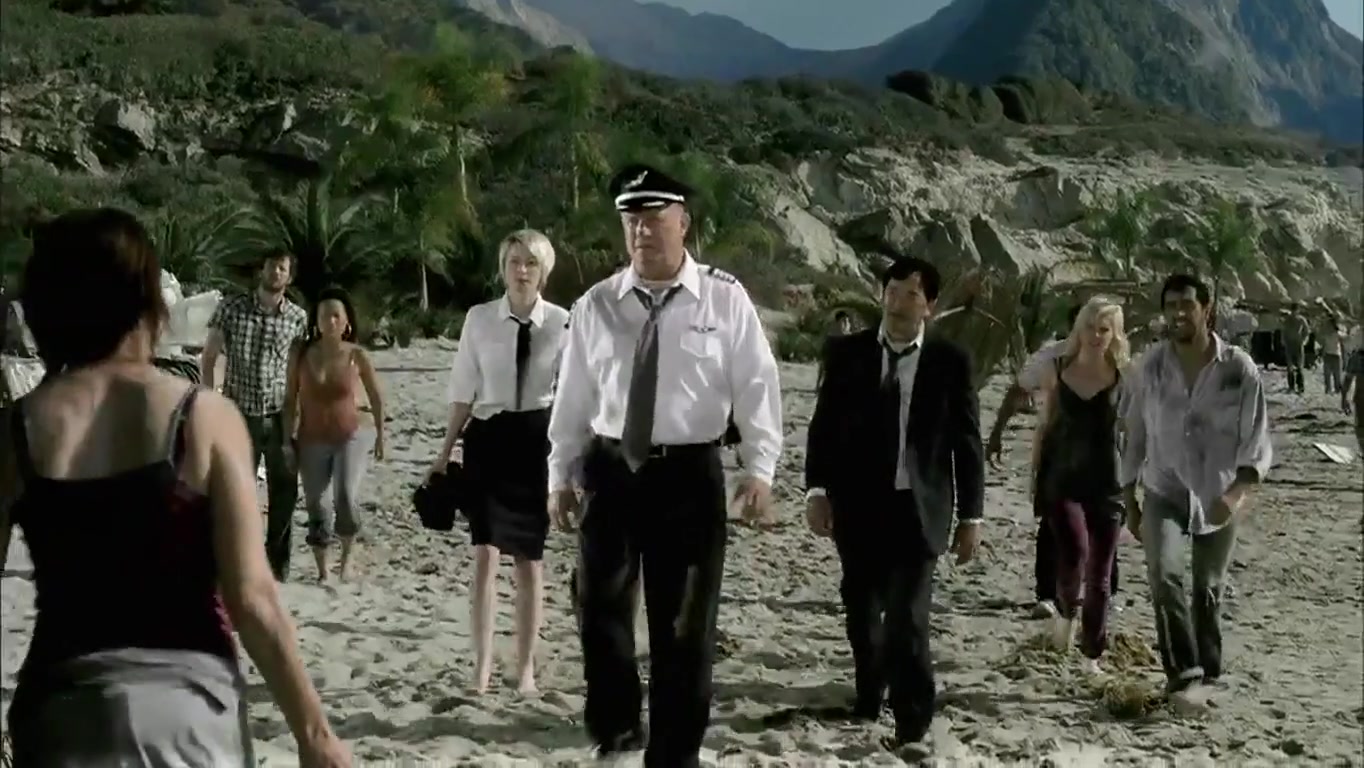}
\vspace{-0.35cm}
\caption{Example ad video frames.}
\label{fig:rawframes}
\vspace{-0.5cm}
\end{figure}

The base channel for our analysis is the raw video frame. The visual content of the raw frames is the exact and complete information available to the viewer when an ad is presented. Theoretically, this representation should carry all the affective information that is meant to be conveyed via the visual modality, however we wanted to explore the utility of this (entire) information for affect recognition. Furthermore, we synthesized a number of visual channels derived from the raw frames, encoding different kinds of information. Figure \ref{fig:rawframes} shows two example frames from an advertisement which serve as references to compare with the synthesized channels.

\subsubsection{Frames Blurred by a Gaussian filter with constant variance (\textbf{Constant Blur})}
\begin{figure}[h]
\centering
\includegraphics[width=0.47\linewidth]{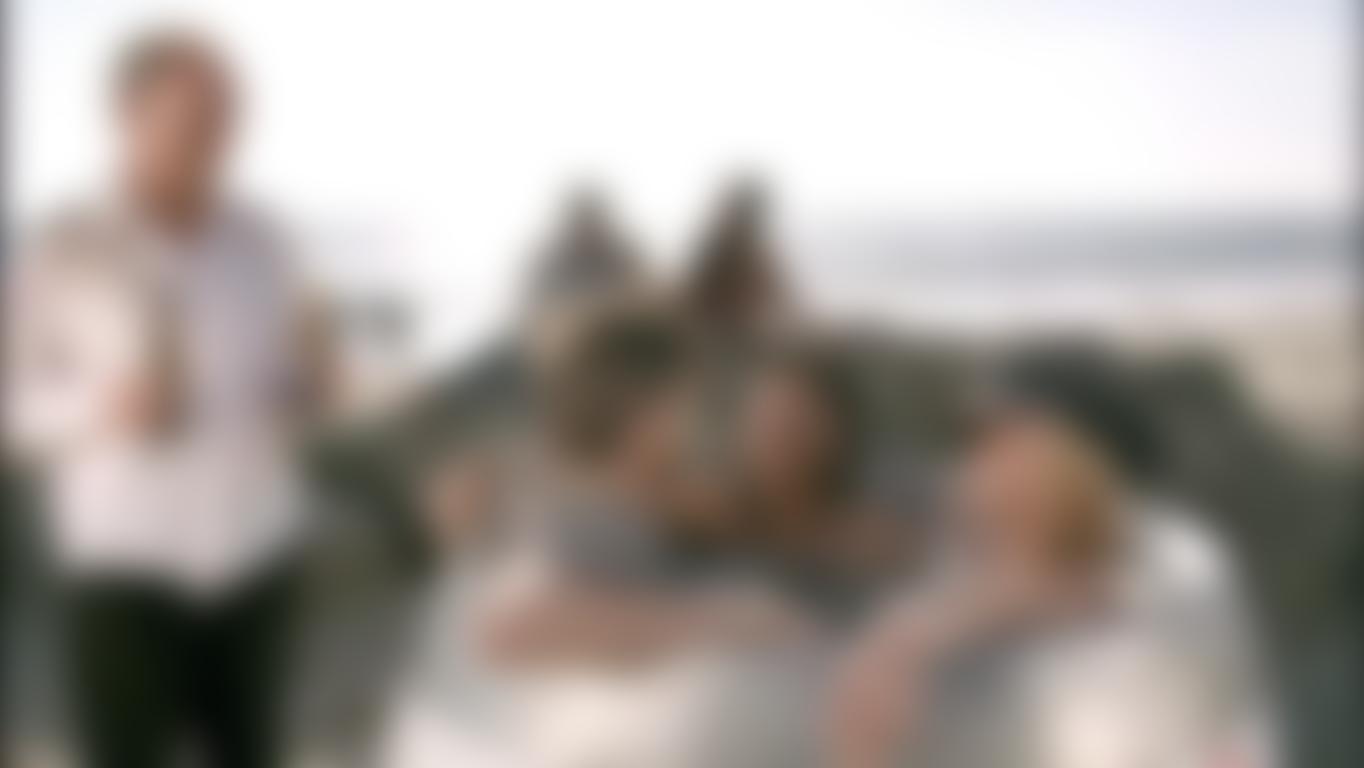}\hfill
\includegraphics[width=0.47\linewidth]{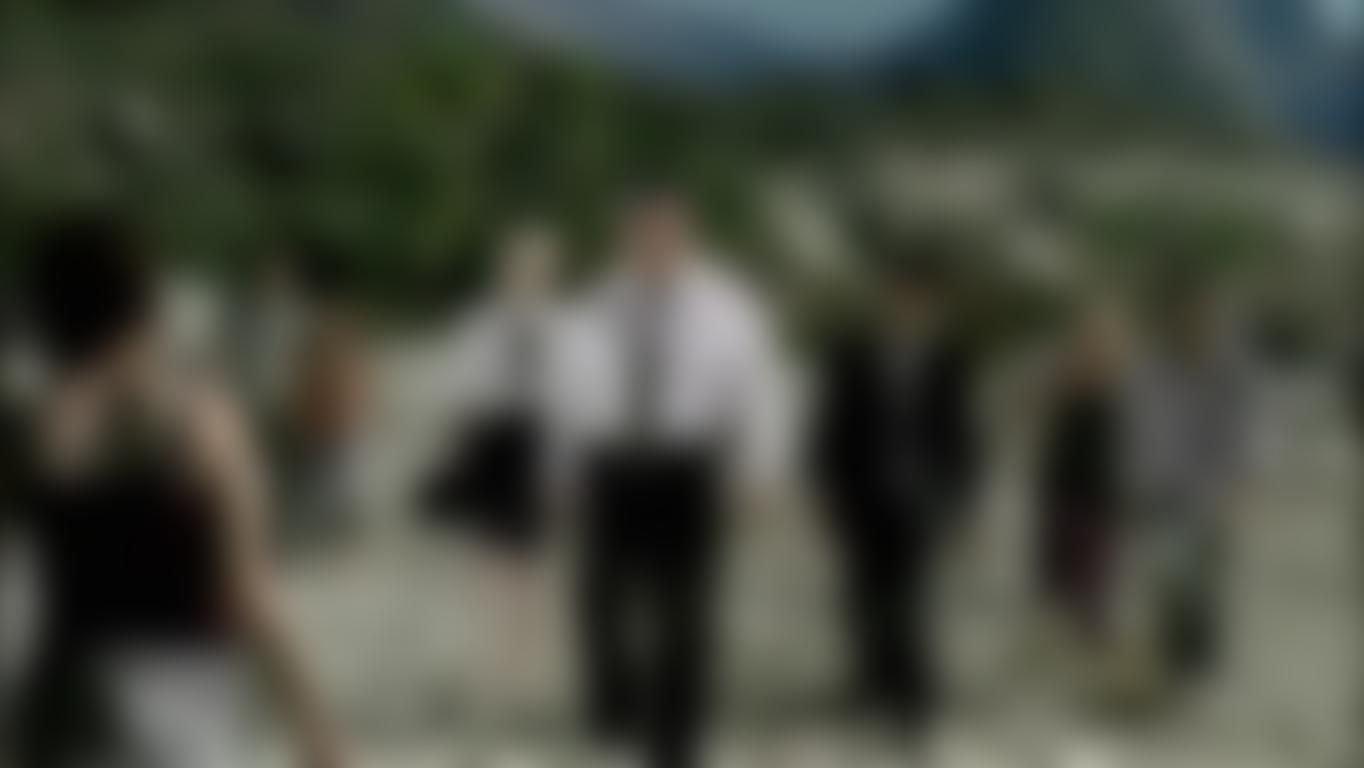}
\vspace{-0.35cm}
\caption{Example images for the \textit{Constant Blur} channel}
\label{fig:constantblur}
\vspace{-0.5cm}
\end{figure}

Prior studies have observed that scene emotion~\cite{Sub14} and memorability~\cite{memorability1} are more correlated with the scene gist rather than finer details.
In order to create a channel that mainly encodes coarse grained scene structure while removing finer details, \textit{i.e.}, to primarily retain the scene \textit{gist}, \textit{ambience} or \textit{setting}, we performed Gaussian blurring on the raw frames with a filter of width $\sigma = 0.2w$, where $w$ denotes video frame width. The resultant images lose many of the sharp details while retaining the coarse scene structure, which we hypothesize contains significant affective information. Figure \ref{fig:constantblur} shows two exemplar frames corresponding to the \textit{Constant Blur} channel for the raw frames in Fig.~\ref{fig:rawframes}.

\subsubsection{Frames Repeatedly Gaussian Blurred until no objects are visually discernible (\textbf{Adaptive Blur})}
This channel is synthesized so as to have no visually discernible objects to the human eye (so that only a `very coarse' scene structure is retained). To synthesize this channel, we use a state-of-the-art object detector. The YOLOv3 \cite{yolov3} object detector is employed to automatically locate salient scene objects belonging to one of the 80 object categories of the MS-COCO \cite{lin2014microsoft} dataset. We run YOLOv3 on the frames on which  \textit{Constant Blur} has been applied, and examine if objects are detected with a confidence threshold of greater than 0.25 (which is the default threshold of YOLOv3). We then repeatedly blur the video frames with the same parameters as in \textit{Constant Blur} until 
no scene object is detected. 

\begin{figure}[h]
\centering
\includegraphics[width=0.47\linewidth]{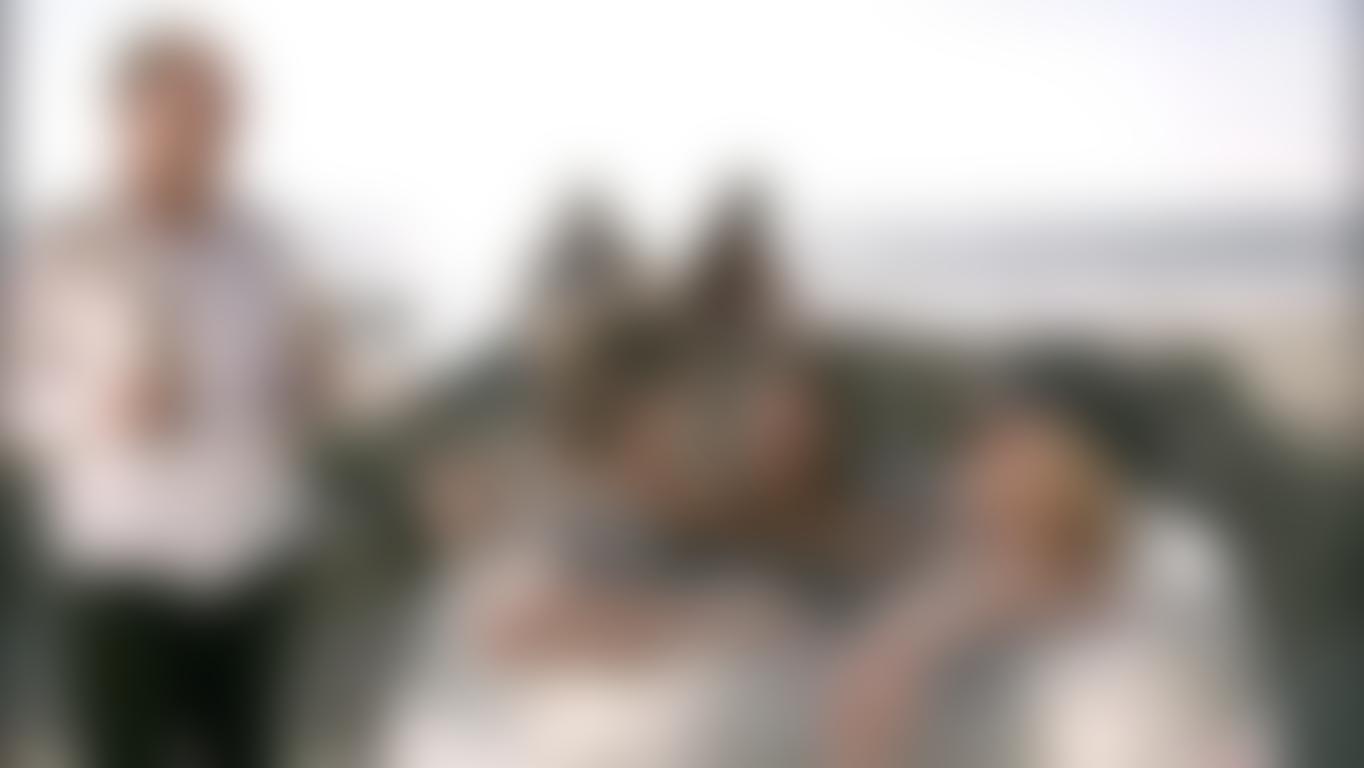}\hfill
\includegraphics[width=0.47\linewidth]{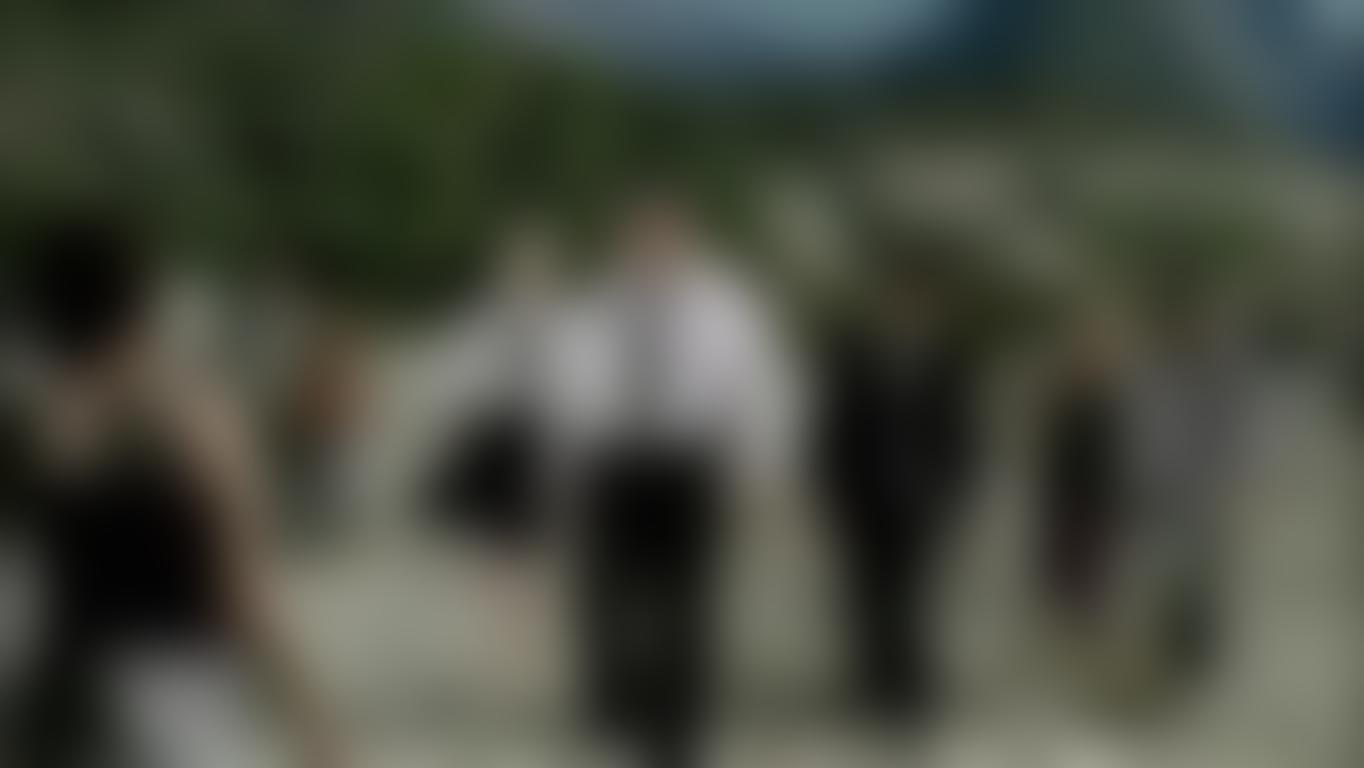}
\vspace{-0.35cm}
\caption{Example images for the \textit{Adaptive Blur} channel}
\label{fig:adaptiveblur}
\vspace{-0.5cm}
\end{figure}

In practice, we needed no more than three repeated blurs with the chosen parameters to ensure zero object detections in any of the 1793 examined frames. Figure \ref{fig:adaptiveblur} illustrates the output of a second blurring on the frames shown in Figure~\ref{fig:constantblur}. In both cases, one person was detected on the left side of the image after the first stage of blurring but no detections following the second blur. Differences between the two pairs of images can be best perceived under zoom. 

\subsubsection{Crops of salient objects detected via YOLOv3 (\textbf{Object Crops})}
\begin{figure}[h]
\centering
\includegraphics[width=0.61\linewidth]{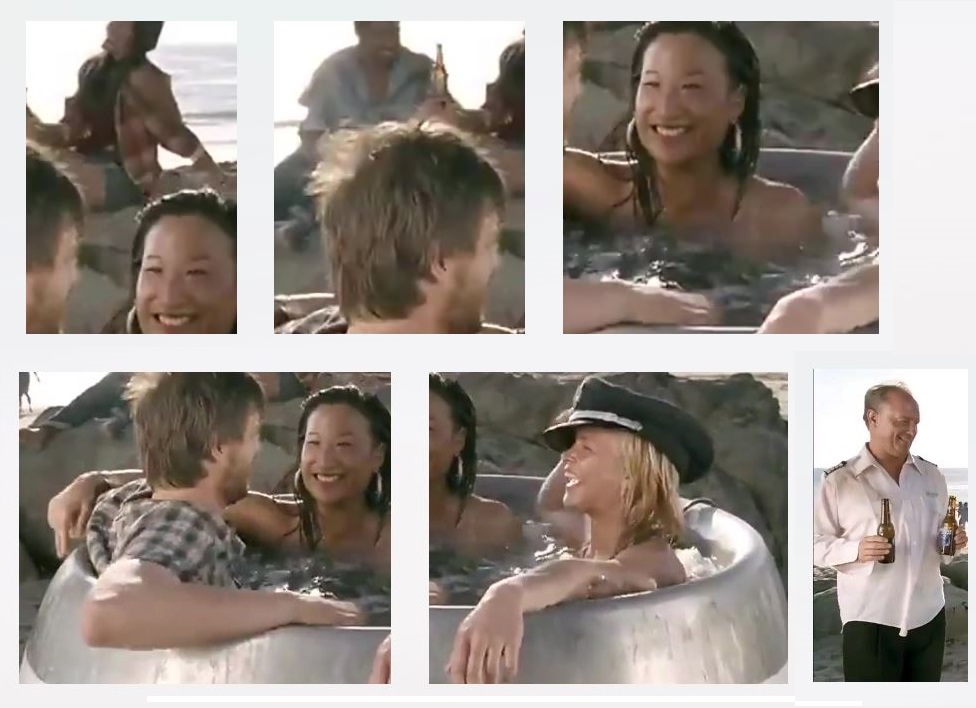}\hfill
\includegraphics[width=0.37\linewidth]{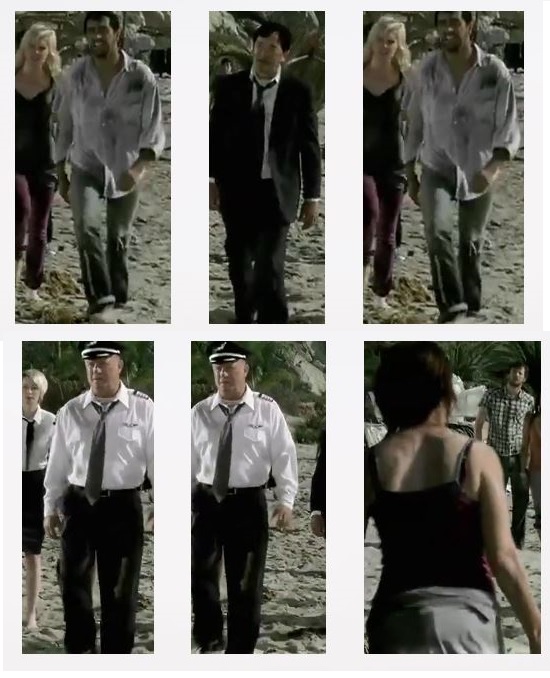}
\vspace{-0.35cm}
\caption{Example outputs for the \textit{Object Crops} channel. Each object crop is fed independently to the affect recognition system.}
\label{fig:objectcrops}
\vspace{-0.5cm}
\end{figure}

A complementary argument to the scene gist influencing its emotion is that the scene affect is a function of the constituent objects and their interactions~\cite{Ramanathan2009}. To measure the contribution of individual scene objects towards the scene affect, we attempted affect prediction from purely object information. To this end, upon running YOLOv3 \cite{yolov3} to perform object detection on the raw frames, we obtained bounding boxes of objects belonging to the 80 MS-COCO \cite{lin2014microsoft} categories (in some instances, we get no detections. \eg, credit or text-only screens or those with uncommon object classes). We then cropped these bounding boxes, and fed them independently as inputs  to the affect recognition model. Figure \ref{fig:objectcrops} shows a set of \textit{Object Crops} for the two reference examples.

\subsubsection{Removal of non-object regions from the frames (\textbf{Object Retained})}
As an alternative to \textit{Object Crops}, where object information is supplied individually to the affect recognition model, we synthesized another channel where the video frame is composed of multiple (detectable) objects. 
To this end, we performed object detection as with \textit{Object Crops}, but instead of using each crop as a separate sample, we blackened (set to 0) those video frame pixels that do not correspond to object detections. This process essentially removes all non-object regions from the video frames.

\subsubsection{ImageNet Softmax Posterior Probabilites (\textbf{AlexNet FC8})}
The two prior channels are limited in the event of spurious detections or no detections at all. In order to have a channel that encodes object related information and also serve as a generic visual descriptor, we used the FC8 layer feature vector of an AlexNet \cite{alex12} network trained on the ImageNet dataset. The FC8 feature vector is 1000 dimensional with each dimension corresponding to the softmax posterior probability of the input video frame containing a particular ImageNet class (out of a 1000 possible classes). We extracted these \textit{AlexNet FC8} features from the raw frames and employed them for affect recognition. The salient aspect of the FC8 features is that they encode object-specific information as well as context in the form of co-occurrence. 

\subsubsection{Gist Features (\textbf{Gist})} We employed Gist features as defined in \cite{oliva2001modeling} as a baseline. These are a classical set of perceptual dimensions (naturalness, openness, roughness, expansion, ruggedness) that represent the dominant spatial structure of a scene. These dimensions may be reliably estimated using spectral and coarsely localized information. The model generates a multidimensional space in which scenes sharing membership in semantic categories (\eg, streets, highways, coasts) are projected close together. We hypothesized a relationship between these dimensions and the affective dimensions of valence and arousal. We compute the Gist descriptor using 4 blocks and 8 orientations per scale.

\subsubsection{Removal of non-gazed regions from video frames (\textbf{Eye ROI})}
\begin{figure}[h]
\centering
\includegraphics[width=0.47\linewidth]{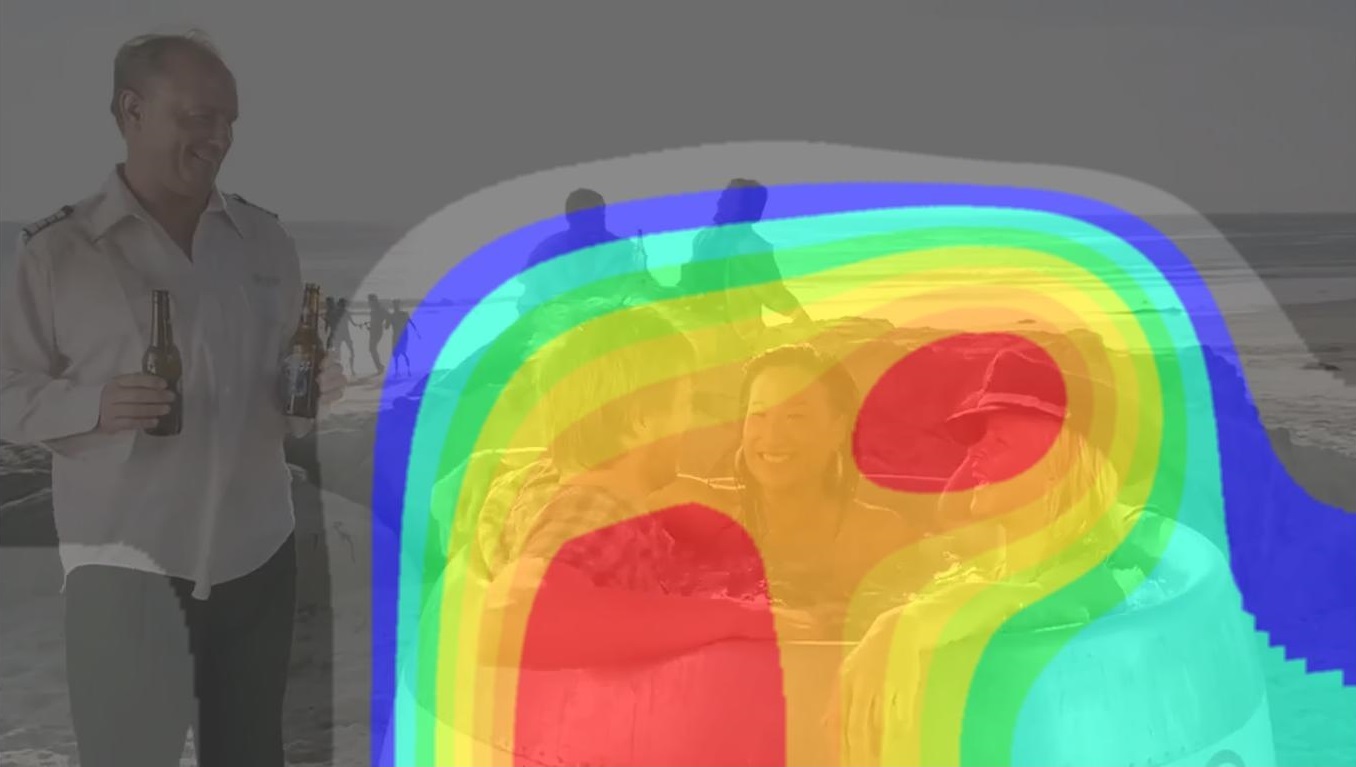}\hfill
\includegraphics[width=0.47\linewidth]{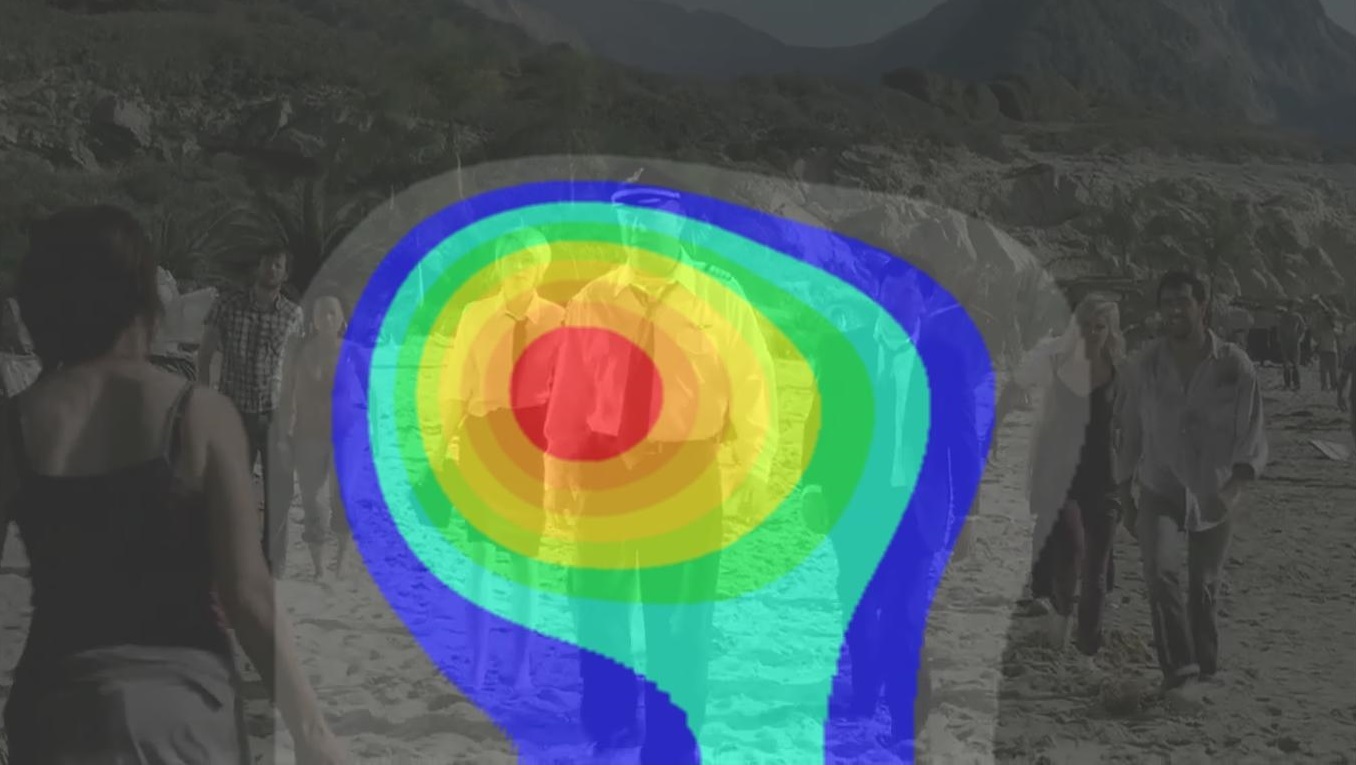}
\vspace{-0.35cm}
\caption{Example eye gaze heatmaps, with warmer colours representing higher gazing frequency (or highly fixated image regions).}
\label{fig:heatmaps}
\vspace{-0.35cm}
\end{figure}

As discussed in Section \ref{eyetrackingprotocol}, eye tracking data for multiple annotators for users who viewed the ads was recorded. From the  eye gaze data, we extracted heatmaps per video frame using the EyeMMV toolbox \cite{krassanakis2014eyemmv}. For the heatmap generation, we used an interval density/spacing of 200 pixels and a Gaussian kernel of size 5 and $\sigma=3$. We normalized gaze frequencies to values between 0 and 255 where 255 represents the image region that was maximally gazed. For each video frame, the gaze points are extracted from a time window of two seconds 
(from one second prior to frame onset to one second post frame presentation). Figure \ref{fig:heatmaps} shows the gaze heatmaps for the two example reference frames.

\begin{figure}[h]
\centering
\includegraphics[width=0.47\linewidth]{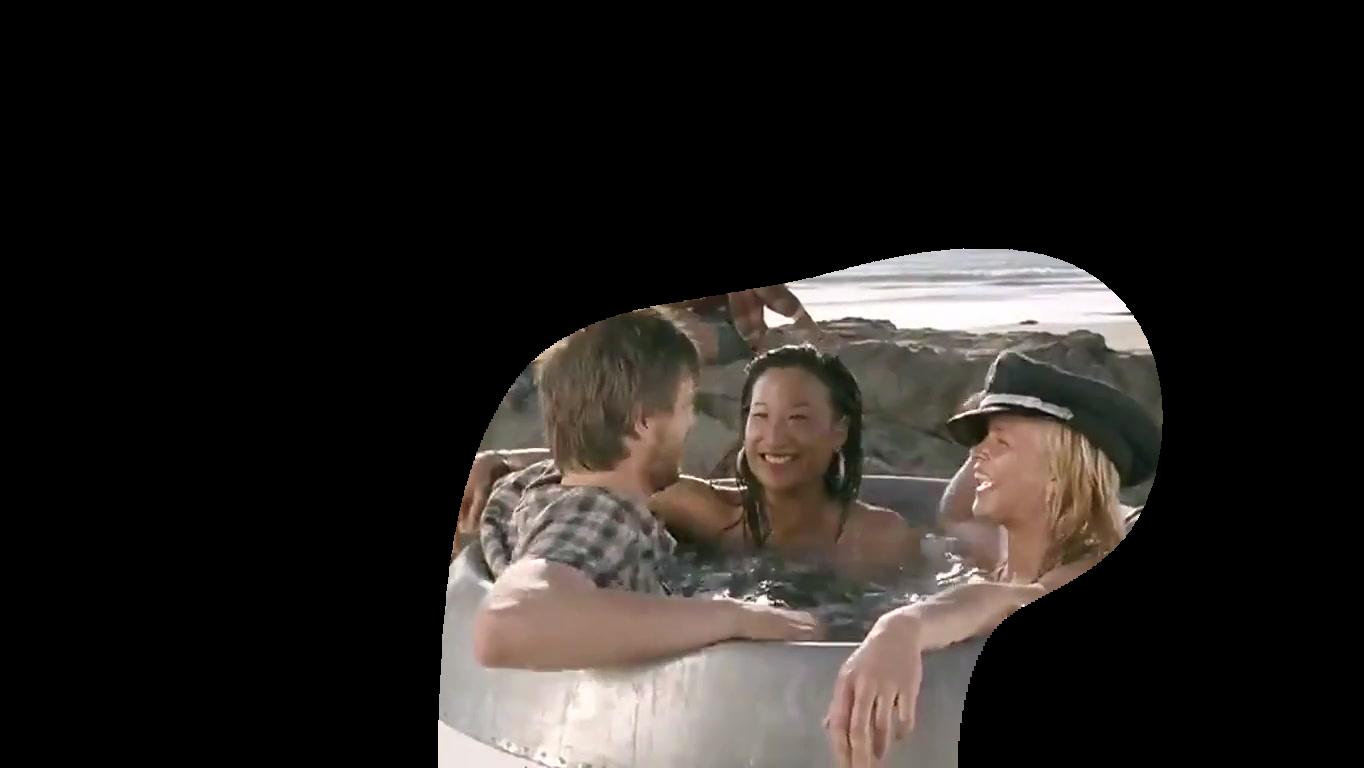}\hfill
\includegraphics[width=0.47\linewidth]{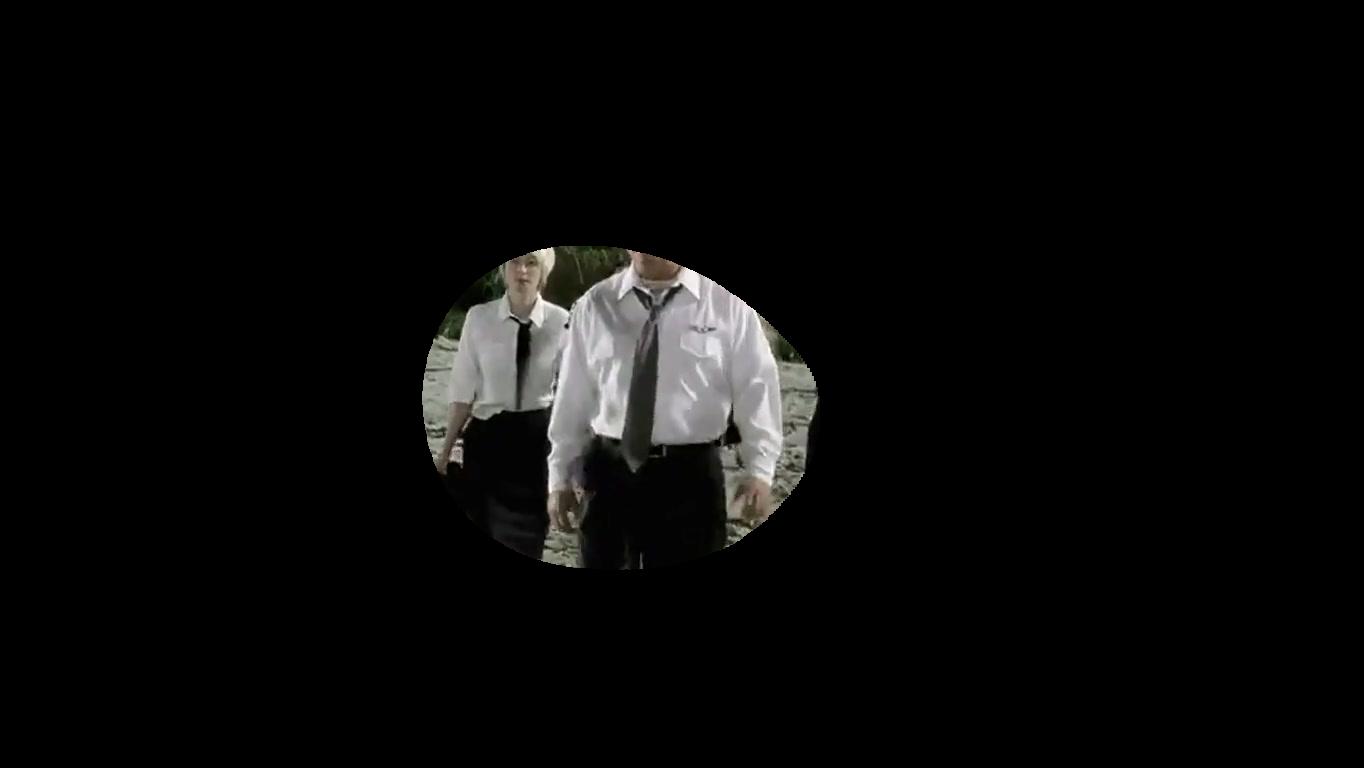}
\vspace{-0.35cm}
\caption{Example images for the \textit{Eye ROI} channel}
\label{fig:eyeroi}
\vspace{-0.5cm}
\end{figure}

Based on these gaze heatmaps, we determined the actively attended image regions. To specifically zero in on the most fixated upon region, we considered only the warmer pixels in the heatmap (yellow and red regions). This channel is similar to the \textit{Objects Retained} channel, but the area to be retained here is determined by human gaze. Similarly as with \textit{objects retained}, we blackened all non-fixated pixels from the video frames. Figure \ref{fig:eyeroi} shows two example outputs for this channel.

\subsubsection{Combination of Gaze Regions with Blurred Context (\textbf{Eye ROI + Context Blur})}
\begin{figure}[h]
\centering
\includegraphics[width=0.47\linewidth]{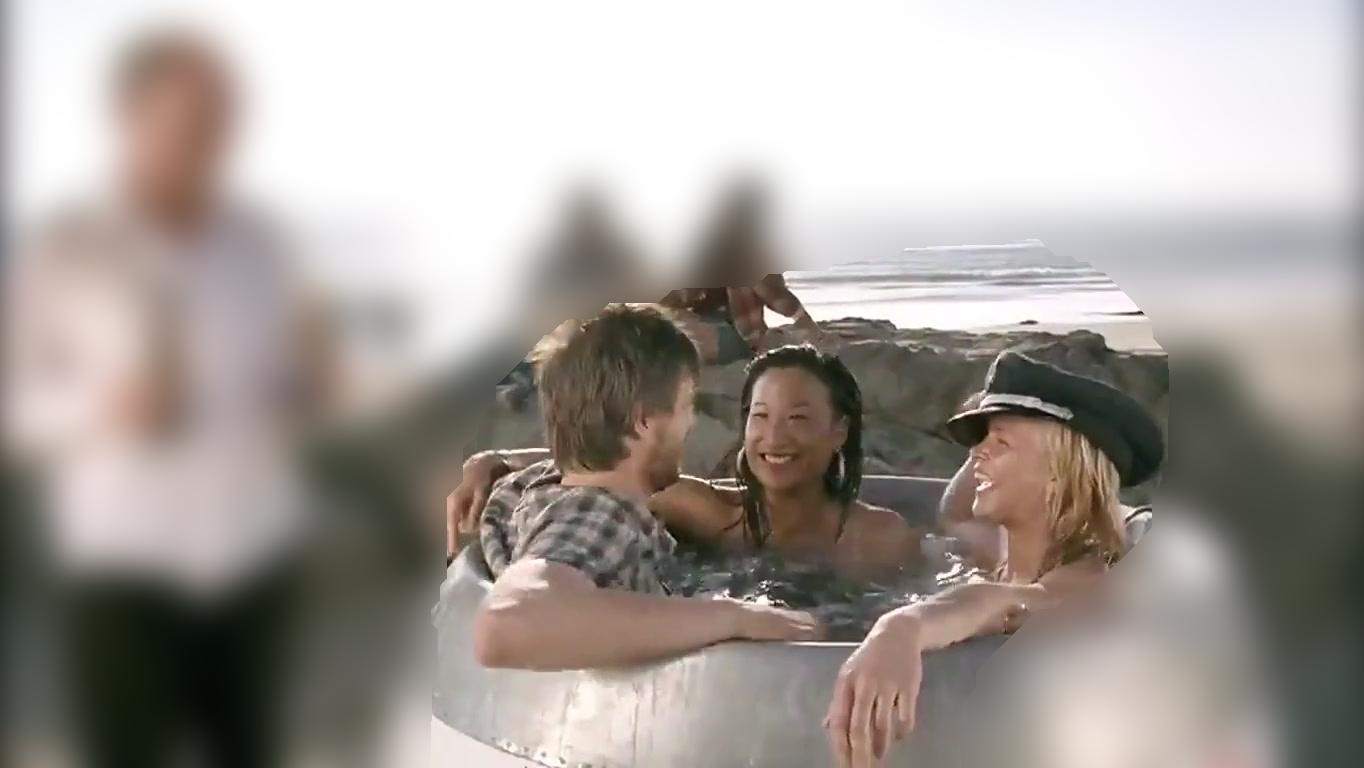}\hfill
\includegraphics[width=0.47\linewidth]{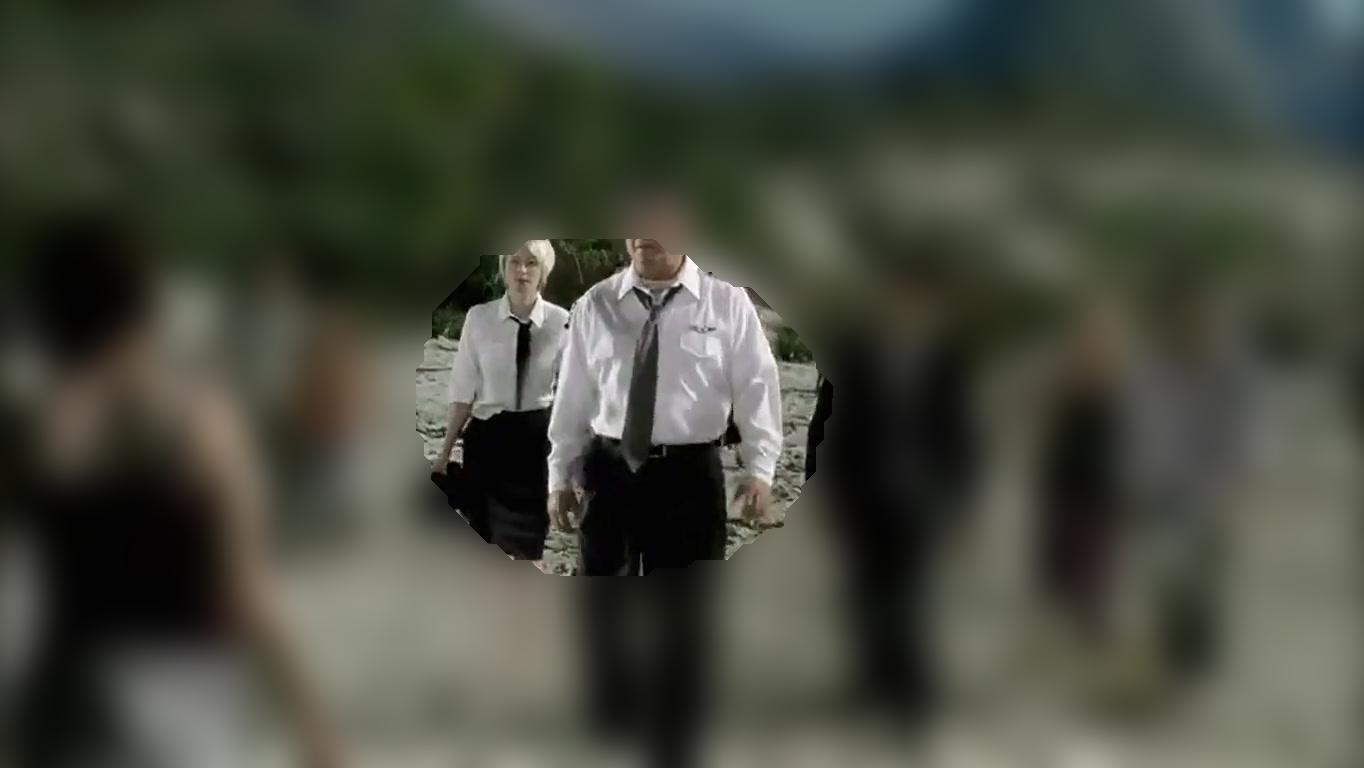}
\vspace{-0.35cm}
\caption{Example images for the \textit{Eye ROI + Context Blur} channel}
\label{fig:eyeroictxblur}
\vspace{-0.5cm}
\end{figure}

In this channel, we wanted to combine the human cues from eye gaze along with the coarse grained contextual information provided by the \textit{Constant Blur} channel. The hypothesis behind this channel being useful for affect prediction is the following. This channel carries both contextual information without the 'noise' induced by finer visual details from the non-salient image regions, and also preserves the most relevant/attended image region as determined by the gaze data. In essence, this channel can be interpreted as a combination of the \textit{Constant Blur} and the \textit{Eye ROI} channels. We simply take the image from the \textit{Constant Blur} channel and then copy the non-zero pixels from the \textit{Eye ROI} channel to the corresponding locations. This yields a 'focus' image in which the salient region is visible clearly, but the surrounding context being encoded at low resolution (similar to foveated rendering in virtual reality applications) . Figure \ref{fig:eyeroictxblur} presents exemplar outputs for this channel.

\subsubsection{Eye Tracking Histogram Features (\textbf{Eye  Hist})}
We also considered purely eye gaze-based statistics as features for affect recognition, as proposed by~\cite{Tavakoli15}. Eye movements are composed of fixations and saccades. A fixation is defined as continuously gazing at a particular location for a specified minimum duration, while a saccade denotes the transition from one fixation to another. Using the EyeMMV toolbox~\cite{krassanakis2014eyemmv}, we extracted fixations with a threshold duration of 100{ms}. From these fixations, we extracted various histograms for quantities like saccade length (50 bin histogram spaced between minimum and maximum lengths), saccade slope (30 bin histogram spaced between minimum and maximum slopes), saccade duration (60 bin histogram spaced between minimum and maximum durations), saccade velocity (50 bin histogram spaced between minimum and maximum velocities), saccade orientation (36 bin histogram spaced at 10 degrees), fixation duration (60 bin histogram spaced between minimum and maximum durations) and spatial fixations (with patches/bins of size 20x40 spread over the whole screen size). We then concatenated each of these features (per annotator) to form a single feature vector for each video. We then used this feature vector characterizing eye movement histograms for affect recognition.

\subsection{Feature Extraction from Image Channels}
This section describes the datasets and methods to create affect recognition models from the different visual channels.
\subsubsection{Datasets}
Due to subjective variance in emotion perception, careful affective labeling is imperative for effectively learning content and user-centric affective correlates~\cite{Hanjalic2005,wang2006affective}, which is why we analyze ads that evoked perfect consensus among experts. Of late, convolutional neural networks (CNNs) have become extremely popular for visual~\cite{alex12} and audio~\cite{Huang2014} recognition, but these models require huge training data. Given the small size of our ad dataset, we fine-tune the pre-trained \textit{Places205}~\cite{places14} model using the affective LIRIS-ACCEDE movie dataset~\cite{baveye2015liris}, and employ this fine-tuned model for encoding ad emotions-- a process known as \textit{\textbf{domain adaptation}} or \textit{\textbf{transfer learning}} in machine learning literature.

\begin{sloppypar}
To learn deep visual features characterizing ad affect, we employed the \textit{Places205} CNN~\cite{places14} intended for image classification. \textit{Places205} is trained using the {Places-205} dataset comprising 2.5 million images and 205 scene categories. The {Places-205} dataset contains a wide variety of scenes with varying illumination, viewpoint and field of view, and we expected a strong correlation between scene perspective, lighting and the scene mood. \textbf{LIRIS-ACCEDE} contains arousal, valence ratings for $\approx$ 10s long movie snippets, whereas the ads are about one minute long (ranging from 30--120s).
\end{sloppypar}

\subsubsection{CNN Training}
For every channel (described above) that is extracted from the LIRIS-ACCEDE dataset, we synthesize the corresponding outputs and use this large collection to fine tune \textit{Places205} for the specific channel. For instance, to fine tune for the \textit{Constant Blur} channel we create a large set of \textit{Constant Blur} images from the approximately 10000 videos from LIRIS-ACCEDE, and use these images to fine tune \textit{Places205}. This model is subsequently used to extract FC7 features for the \textit{Constant Blur} images synthesized for the ad dataset. This is not possible for the eye tracking based channels because LIRIS-ACCEDE doesn't have eye tracking based annotations. For channels involving eye tracking data (\textit{Eye ROI, Eye ROI + Context Blur}), we use the model fine tuned on the raw image data for feature extraction.

We use the Caffe~\cite{caffe} deep learning framework for fine-tuning \textit{Places205}, with a momentum of 0.9, weight decay of 0.0005, and a base learning rate of 0.0001 reduced by $\frac{1}{10}^{th}$ every 20000 iterations. We train multiple binary affect prediction networks this way to recognize high and low arousal and valence from FC7 features of each channel (excepting FC8 features and eye movement histograms, for which the features/histograms themselves are employed as features). To fine-tune \textit{Places205}, we use only the top and bottom 1/3rd LIRIS-ACCEDE videos in terms of arousal and valence rankings under the assumption that descriptors learned for the extreme-rated clips will more effectively represent affective concepts than the neutral-rated clips. 4096 dimensional {FC7} features extracted from the penultimate fully connected layer of the fine tuned networks are used for ad affect recognition.

\section{Experiments and Results}~\label{sec:er}
We first provide a brief description of the classifiers used and settings employed for  binary affect prediction (\ie, an ad is predicted as invoking \textit{high} or \textit{low }valence/arousal). Ground truth labels are provided by the experts, and these labels are in agreement with the user ratings as described in Section \ref{sec:dataset}. 

\paragraph*{Classifiers:} We employed the Linear Discriminant Analysis (LDA), linear SVM (LSVM) and Radial Basis SVM (RSVM) classifiers in our affect recognition experiments. LDA and LSVM separate H/L labeled training data with a hyperplane, while RSVM enables non-linear separation between the H and L classes via transformation onto a high-dimensional feature space. We used these three classifiers on input FC7 features from the fine tuned networks described in the previous section. For \textit{AlexNet FC8} \cite{alex12}, we used the FC8 features denoting the output of an object classification network. Similarly, \textit{Eye Hist} features were directly input to the LDA, linear SVM and radial basis SVM classifiers.

\paragraph*{Metrics and Experimental Settings:} We used the F1-score (F1), defined as the harmonic mean of precision and recall as our performance metric, due to the unbalanced distribution of positive and negative samples. As we evaluate affect recognition performance on a small dataset, affect recognition results obtained over 10 repetitions of 5-fold cross validation (CV) (total of 50 runs) are presented. CV is typically used to overcome the \textit{overfitting} problem on small datasets, and the optimal SVM parameters are determined from the range $[10^{-3},10^{3}]$ via an inner five-fold CV on the training set. Finally, in order to examine the temporal variance in affect recognition performance, we present F1-scores obtained over (a) all ad frames (`All'), (b) last 30s (L30) and (c) last 10s (L10) for affect recognition from all visual modalities.

\begin{table*}[htbp]
\fontsize{8}{8}\selectfont
\renewcommand{\arraystretch}{1.3}
\centering
\caption{Advertisement Affect Recognition. F1 scores are presented in the form $\mu \pm \sigma$. } \label{tab:cap} \vspace{-.2cm}
\begin{tabular}{|c|ccc|ccc|}
  \hline
 
	\multicolumn{1}{|c|}{\textbf{Method}} & \multicolumn{3}{c|}{\textbf{valence}} & \multicolumn{3}{c|}{\textbf{Arousal}} \\ \hline 
	\multicolumn{1}{|c|}{~} & {\textbf{F1 (all)}} & {\textbf{F1 (L30)}} & {\textbf{F1 (L10)}}  & {\textbf{F1 (all)}} & {\textbf{F1 (L30)}} & {\textbf{F1 (L10)}}\\ \hline

	
	\textbf{Video FC7 + LDA}  & 0.69$\pm$0.03 & 0.68$\pm$0.11 & 0.60$\pm$0.19 & 0.66$\pm$0.03 & 0.55$\pm$0.09 & {0.42$\pm$0.14}\\
	\textbf{Video FC7 + LSVM} & 0.66$\pm$0.02 & 0.64$\pm$0.10 & 0.57$\pm$0.19 & 0.63$\pm$0.02 & 0.58$\pm$0.10 & 0.47$\pm$0.15\\
	\textbf{Video FC7 + RSVM} & \textbf{0.71$\pm$0.02} & {0.70$\pm$0.10} & {0.61$\pm$0.18} & \textbf{0.68$\pm$0.02} & {0.57$\pm$0.11} & {0.42$\pm$0.14}\\ \hline
    
    \textbf{Constant Blur FC7 + LDA}  & 0.70$\pm$0.02 & 0.67$\pm$0.10 & 0.63$\pm$0.21 & 0.69$\pm$0.03 & 0.59$\pm$0.10 & {0.46$\pm$0.16}\\
	\textbf{Constant Blur FC7 + LSVM} & 0.67$\pm$0.02 & 0.68$\pm$0.10 & 0.62$\pm$0.23 & 0.61$\pm$0.03 & 0.55$\pm$0.11 & 0.45$\pm$0.20\\
	\textbf{Constant Blur FC7 + RSVM} & \textbf{0.73$\pm$0.02} & {0.70$\pm$0.09} & {0.65$\pm$0.21} & \textbf{0.70$\pm$0.02} & {0.59$\pm$0.09} & {0.45$\pm$0.16}\\ \hline
    
        \textbf{Adaptive Blur FC7 + LDA}  & 0.71$\pm$0.02 & 0.67$\pm$0.10 & 0.63$\pm$0.21 & \textbf{0.70$\pm$0.02} & 0.59$\pm$0.11 & {0.47$\pm$0.15}\\
	\textbf{Adaptive Blur FC7 + LSVM} & 0.66$\pm$0.02 & 0.69$\pm$0.10 & 0.62$\pm$0.23 & 0.62$\pm$0.03 & 0.55$\pm$0.11 & 0.46$\pm$0.19\\
	\textbf{Adaptive Blur FC7 + RSVM} & \textbf{0.72$\pm$0.02} & {0.70$\pm$0.09} & {0.64$\pm$0.21} & {0.69$\pm$0.02} & {0.59$\pm$0.09} & {0.48$\pm$0.14}\\ \hline
    
        \textbf{Object Crops FC7 + LDA}  & 0.59$\pm$0.04 & 0.60$\pm$0.10 & 0.54$\pm$0.17 & 0.58$\pm$0.03 & 0.55$\pm$0.07 & {0.42$\pm$0.14}\\
	\textbf{Object Crops FC7 + LSVM} & 0.59$\pm$0.03 & 0.59$\pm$0.10 & 0.54$\pm$0.17 & 0.56$\pm$0.03 & 0.54$\pm$0.10 & 0.45$\pm$0.15\\
	\textbf{Object Crops FC7 + RSVM} & {0.63$\pm$0.04} & \textbf{0.63$\pm$0.10} & {0.58$\pm$0.17} & \textbf{0.61$\pm$0.03} & {0.59$\pm$0.08} & {0.47$\pm$0.15}\\ \hline
    
        \textbf{Object Retained FC7 + LDA}  & 0.59$\pm$0.03 & 0.56$\pm$0.08 & 0.57$\pm$0.21 & 0.55$\pm$0.03 & 0.51$\pm$0.10 & {0.56$\pm$0.20}\\
	\textbf{Object Retained FC7 + LSVM} & 0.57$\pm$0.03 & 0.57$\pm$0.10 & 0.53$\pm$0.20 & 0.56$\pm$0.04 & 0.54$\pm$0.09 & \textbf{0.58$\pm$0.17}\\
	\textbf{Object Retained FC7 + RSVM} & \textbf{0.61$\pm$0.03} & {0.58$\pm$0.09} & {0.56$\pm$0.21} & {0.55$\pm$0.03} & {0.50$\pm$0.11} & {0.55$\pm$0.19}\\ \hline
    
        \textbf{AlexNet FC8 + LDA}  & 0.71$\pm$0.02 & 0.68$\pm$0.11 & 0.71$\pm$0.18 & \textbf{0.67$\pm$0.03} & 0.58$\pm$0.10 & {0.51$\pm$0.15}\\
	\textbf{AlexNet FC8 + LSVM} & 0.70$\pm$0.02 & 0.70$\pm$0.09 & 0.72$\pm$0.19 & 0.66$\pm$0.02 & 0.58$\pm$0.11 & 0.51$\pm$0.17\\
	\textbf{AlexNet FC8 + RSVM} & {0.71$\pm$0.02} & {0.69$\pm$0.09} & \textbf{0.73$\pm$0.17} & {0.66$\pm$0.02} & {0.55$\pm$0.10} & {0.50$\pm$0.17}\\ \hline    
        
        \textbf{Gist + LDA}  & 0.57$\pm$0.03 & 0.52$\pm$0.10 & 0.43$\pm$0.16 & 0.58$\pm$0.02 & 0.52$\pm$0.09 & {0.46$\pm$0.19}\\
	\textbf{Gist + LSVM} & \textbf{0.57$\pm$0.03} & 0.53$\pm$0.10 & 0.42$\pm$0.16 & \textbf{0.58$\pm$0.02} & 0.52$\pm$0.08 & 0.46$\pm$0.17\\
	\textbf{Gist + RSVM} & {0.39$\pm$0.02} & {0.36$\pm$0.06} & {0.35$\pm$0.09} & {0.56$\pm$0.02} & {0.54$\pm$0.09} & {0.52$\pm$0.17}\\ \hline
    
    
        \textbf{Eye ROI FC7 + LDA}  & 0.65$\pm$0.02 & 0.61$\pm$0.10 & 0.60$\pm$0.21 & 0.68$\pm$0.02 & 0.56$\pm$0.09 & {0.53$\pm$0.19}\\
	\textbf{Eye ROI FC7 + LSVM} & 0.62$\pm$0.03 & 0.58$\pm$0.07 & 0.55$\pm$0.21 & 0.66$\pm$0.02 & 0.59$\pm$0.09 & 0.59$\pm$0.18\\
	\textbf{Eye ROI FC7 + RSVM} & \textbf{0.68$\pm$0.02} & {0.63$\pm$0.10} & {0.62$\pm$0.20} & \textbf{0.70$\pm$0.02} & {0.56$\pm$0.08} & {0.54$\pm$0.18}\\ \hline

        \textbf{Eye Hist + LDA}  & 0.53$\pm$0.04 & 0.57$\pm$0.04 & 0.50$\pm$0.05 & 0.52$\pm$0.03 & 0.54$\pm$0.05 & {0.56$\pm$0.02}\\
	\textbf{Eye Hist + LSVM} & 0.52$\pm$0.04 & 0.54$\pm$0.03 & 0.52$\pm$0.05 & 0.53$\pm$0.04 & 0.53$\pm$0.06 & 0.55$\pm$0.04\\
	\textbf{Eye Hist + RSVM} & {0.53$\pm$0.04} & \textbf{0.58$\pm$0.04} & {0.52$\pm$0.05} & {0.52$\pm$0.04} & \textbf{0.56$\pm$0.06} & {0.56$\pm$0.02}\\ \hline
    
    
        \textbf{Eye ROI + Context Blur FC7 + LDA}  & 0.69$\pm$0.02 & 0.71$\pm$0.09 & 0.65$\pm$0.16 & 0.69$\pm$0.02 & 0.61$\pm$0.10 & {0.63$\pm$0.18}\\
	\textbf{Eye ROI + Context Blur FC7 + LSVM} & 0.67$\pm$0.02 & 0.67$\pm$0.11 & 0.59$\pm$0.18 & 0.69$\pm$0.02 & 0.61$\pm$0.09 & 0.67$\pm$0.18\\
	\textbf{Eye ROI + Context Blur FC7 + RSVM} & \textbf{0.74$\pm$0.02} & {0.74$\pm$0.08} & {0.65$\pm$0.19} & \textbf{0.73$\pm$0.02} & {0.64$\pm$0.09} & {0.61$\pm$0.16}\\ \hline

  \hline
\end{tabular} \vspace{-.2cm}
\end{table*}

\subsection{Results Overview}~\label{subsec:resoview}
Table \ref{tab:cap} contains the performance for each method for both valence and arousal recognition in all the experimental settings. The highest F1 score obtained over all the temporal windows are denoted in bold for each method. Based on the performance of the methods, we observe the following.

Focusing on all classifiers that operate on deep learning based features, valence (Peak F1 = 0.74) is recognized generally better than arousal (Peak F1 = 0.73). The only two methods for which peak arousal recognition is better than valence are the \textit{Gist} features (Peak Arousal F1 = 0.58, Peak valence F1 = 0.57) and the \textit{Eye ROI} method (Peak valence F1 = 0.68, Peak Arousal F1 = 0.70). This goes on to show that visual information based modalities contain more information related to valence than arousal, which is consistent with expectation and as shown in prior works \cite{Shukla2017acm,Shukla2017icmi}.

Relatively small F1-score variances are noted for the 'all' condition with the adopted cross validation procedure in Table~\ref{tab:cap}, especially with fc7 features indicating that the affect recognition results are {minimally impacted} by  overfitting. The $\sigma$ values for the 'L30' and the 'L10' settings are uniformly and progressively greater for all methods involving deep features. There can also be seen a general degradation in F1 score over the three settings. Affect recognition performance being lower in the L30 and L10 conditions highlights both a lack of heterogeneity in affective information towards the end of the ad and the limitation of using a \textit{single} affective label (as opposed to continuous labeling, \eg, using \textit{Feeltrace}~ \cite{cowie2000feeltrace}) over time. Generally lower F1-scores achieved for arousal with all methods suggests that arousal is more difficult to model than valence, and that coherency between visual features and valence labels is sustainable over time.

The deep learning based FC7/FC8 features, even while not being interpretable, considerably outperform the handcrafted \textit{Gist} and the \textit{Eye Hist} features. This can be attributed to a variety of reasons. Even though Tavakoli \etal~\cite{Tavakoli15} show that eye tracking features correlate well with valence, their approach is validated on images whereas we focus on ad videos which have different eye gaze dynamics, influenced by engaging ad content. The coarse visual scene structure (\textit{Constant} and \textit{Adaptive Blur} conditions) in itself proves to be considerably more informative for affect prediction than fixation statistics. Similarly, while gist features may be able to effectively encode characteristics of simple scenes (\eg, indoor vs outdoor), ad videos may contain a wide variety in scene content (\eg, such as a sudden transformation from indoor to outdoor), which renders them ineffective for describing visual ad content.

Comparing the various channels, we notice a sharp difference in performance between channels that only encode object-related information vs. those that encode contextual information. The best performing computationally driven object channel is \textit{Object Crops} (Peak valence F1 = 0.63, Peak Arousal F1 = 0.61), which performs significantly worse than the best performing contextual channel \textit{Constant Blur} (Peak valence F1 = 0.73, Peak Arousal F1 = 0.70). The \textit{Alexnet FC8} channel which contains information regarding the existence and co-occurrence of 1000 object classes in a video frame as compared to the YOLOv3 detector that only handles 80 classes, performs much better than the \textit{Object Crops} and \textit{Objects Retained} channels. Among object based channels, \textit{Alexnet FC8} performs the best and only performs marginally worse than the context-based blur channels for valence recognition. However, the blurred video channels appear to encode information more pertinent for arousal recognition as compared to \textit{Alexnet FC8}.

Interestingly, we also observe that the performance of the \textit{Constant Blur} and the \textit{Adaptive Blur} channels is slightly better than the performance of the \textit{Video} channel (Peak valence F1 = 0.71, Peak Arousal F1 = 0.68) containing the complete video information. Their performances are very comparable, which is largely due to the fact that the two channels encode largely similar information (for many video frames, no objects could be detected after the first blurring step). The \textit{Eye ROI} channel clearly outperforms the \textit{Objects Retained} channel, which points to human driven salient regions containing more affective information than computationally detected objects.

Lastly, the best arousal and valence recognition is achieved with the \textit{Eye ROI + Context Blur} channel (Peak valence F1 = 0.74, Peak Arousal F1 = 0.73) which encodes high resolution information corresponding to the gazed (or interesting) content, along with the coarse grained scene context (as the rest of the scene is blurred). This observation implies that eliminating `noise' in the form of finer details pertaining to the scene context is beneficial for affect recognition. 
Context noise is also known to affect (low-level) visual tasks such as object classification-- \eg, noise arising from partial matches of high resolution background information has been shown to confound object categorization in~\cite{katti2016object}.

\section{Discussion}~\label{subsec:disc}

A long-standing debate exists regarding what influences emotions while viewing visual content-- individual scene objects and their interactions central to the narrative, or the overall scene structure (that we call the gist or context). Particularly with reference to ad design and creation, the ad director has to decide whether to a) focus on key elements conveying the ad emotion, or b) the context surrounding these elements as well. 
In this regard, the findings of this study provide some interesting takeaways. Our results suggest that both the central scene objects (captured by user gaze and encoded in high resolution) and the scene context play a critical role towards conveying the scene affect. 

The most striking result that we have obtained is that coarse-grained scene structure conveys affect better than object-specific information, eye-gaze related measures and even the \textit{original video}. Although we know that coarse scene structure can influence scene description~\cite{feifei2002} and object perception~\cite{ Bar2004} irrespective of the presence ~\cite{ Bar2004,katti2017targets} or absence of target objects~\cite{ katti2017targets,katti2016object}, it is indeed surprising that coarse scene descriptors can effectively predict valence and arousal in video advertisements. Also, the obtained results point to the importance of capturing user-specific information (via gaze tracking), and highlight the limitations of content descriptors (including deep features) for affect recognition.

How are our findings useful in the context of ad affect recognition and computational advertising, where the objective is to insert the most appropriate ads at optimal locations within a streamed video? Past work has shown that such affective information can significantly influence the evaluation~\cite{Truss132} and impact of consumer behavior~\cite{Broach1995}.
In terms of affect mining, we find that affective information can be reliably modeled even from short ad segments, given the reasonable recognition achieved with L30 and L10 ad segments. 

Also, emotions invoked by ads can be identified well using only primitive or blurred scene descriptors, which suggest that a computational advertising framework can be developed with relatively low complexity. Overall, observed results translate into the following ad design principles (a) Target valence and arousal should preferably be conveyed by the scene background in addition to being embedded in the ad narrative or object information. (b) The fact that models trained with information from all detectable objects do not do as well as ones trained with few actively attended objects suggests that cluttered scenes can hamper affective processing and should be avoided. 

Can we leverage our insights to achieve more effective computational advertising? Content based features extracted using image statistics ~\cite{cavva} and deep neural network representations ~\cite{Shukla2017icmi,Shukla2017acm} have been used for affect-driven placement of advertisements within streamed video. Our work suggests that optimal ad-video scene matching at the content plus emotional levels can be achieved by using traditional object annotation methods for content matching, and a coarse scene structure matching for emotional indexing. From a methods perspective, embedding high resolution objects into coarse backgrounds is compatible with existing deep neural architectures, and represents an easily implementable pipeline. 

\section{Conclusions and Future Work}~\label{sec:cfw}
Analyses of video-based information channels confirm that interesting scene objects, captured via eye gaze and encoded at high resolution, along with coarse information regarding the remaining scene structure (or context) best capture the scene emotion. The different channels explored in this work can be synthesized with minimal human intervention (only eye tracking information is required), and the proposed affect recognition models are highly scalable owing to the nature of computer vision and machine learning models used. These channels can be used to annotate large video datasets for a variety of scene understanding applications.

We aim to fully automate our analytical pipeline in future by replacing human eye-gaze with state-of-art visual saliency prediction algorithms. The proposed ad affect recognition models can be combined with a computational advertising framework like {CAVVA}~\cite{cavva} to maximize engagement, immediate and long term brand recall and click through rates as validated by previous work \cite{Shukla2017acm,Shukla2017icmi}.
\begin{sloppypar}
In future, we would also like to explore decompositions of information in the audio, speech and text modalities and how they can be useful for affect prediction. Another interesting line for future research is the development of a system that learns and recommends impactful scene structures for the creation  of emotional advertisement videos.
\end{sloppypar}

\begin{acks}
This research is supported by the National Research
Foundation, Prime Ministers Office, Singapore under its
International Research Centre in Singapore Funding Initiative.
\end{acks}

\bibliographystyle{ACM-Reference-Format}
\balance
\bibliography{context_gaze_affect_ads}

\end{document}